# Untargeted, Targeted and Universal Adversarial Attacks and Defenses on Time Series

Pradeep Rathore[1], Arghya Basak[1], Sri Harsha Nistala, Venkataramana Runkana
TCS Research
Pune, India
Email: {rathore.pradeep, arghya.basak, sriharsha.nistala, venkat.runkana}@tcs.com

*Abstract*—Deep learning based models are vulnerable to adversarial attacks. These attacks can be much more harmful in case of targeted attacks, where an attacker tries not only to fool the deep learning model, but also to misguide the model to predict a specific class. Such targeted and untargeted attacks are specifically tailored for an individual sample and require addition of an imperceptible noise to the sample. In contrast, universal adversarial attack calculates a special imperceptible noise which can be added to any sample of the given dataset so that, the deep learning model is forced to predict a wrong class. To the best of our knowledge these targeted and universal attacks on time series data have not been studied in any of the previous works. In this work, we have performed untargeted, targeted and universal adversarial attacks on UCR time series datasets. Our results show that deep learning based time series classification models are vulnerable to these attacks. We also show that universal adversarial attacks have good generalization property as it need only a fraction of the training data. We have also performed adversarial training based adversarial defense. Our results show that models trained adversarially using Fast gradient sign method (FGSM), a single step attack, are able to defend against FGSM as well as Basic iterative method (BIM), a popular iterative attack.

I. INTRODUCTION

Deep learning has shown great success in real life applications based on computer vision, natural language processing, speech recognition and time series modelling. Deep learning based models are used for many time series problems like disease detection and classification using ECG signals [1], anomaly detection and diagnosis using sensory data from industries such as iron and steel, oil and gas, fine chemicals, thermal and nuclear power plant [2,3], etc. These deep learning based systems are vulnerable to adversarial attacks. These attacks are easy to perform and can lead to serious security implications.

Adversarial attacks were first studied by Szedegy et al. [4] who performed adversarial attacks on images and successfully fooled deep learning models by adding an imperceptible noise to the input. Adversarial attacks can be classified according to adversarial goals, adversarial knowledge and perturbation scope [5]. The taxonomy is described below :

Adversarial goals

- Targeted attacks are the attacks where the attacker tries to misguide the model to a particular class other than the true class.

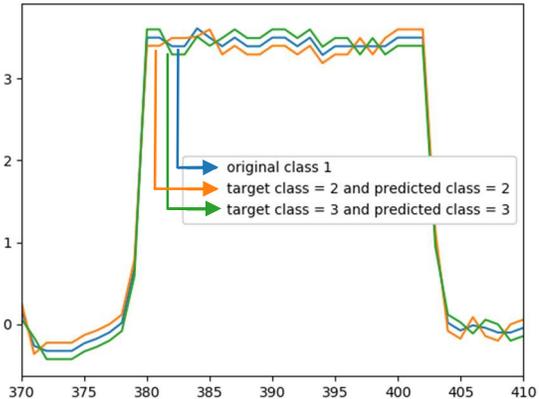

Fig. 1 Example of targeted adversarial attack on 'Large Kitchen Appliances' dataset by adding imperceptible noise

- Untargeted attacks are the attacks where the attacker tries to misguide the model to predict any of the incorrect classes.

Adversarial knowledge

- White-box attacks assume that the attacker knows everything about the model's architecture, parameters, training dataset etc.
- Black-box attacks assume that the attacker can query the model but does not have access to the model's architecture, parameters, training dataset etc.

Perturbation scope

- Individual attacks are done by solving an optimization problem for each sample. Adversarial perturbation for each sample is different.
- Universal attacks are done by solving an optimization problem for the whole dataset. Addition of the same noise to different input samples can output wrong class for each of the input samples.

One of the most dangerous examples of adversarial attack is altering the output of deep learning models used in autonomous vehicles to visually detect traffic signals. Eykholt et al. [6] successfully performed targeted attacks on traffic signal images. Kurakin et al. [7] performed black-box adversarial attacks using printed physical adversarial example images to fool tensorflow demo app object classifier on a mobile phone. Fawaz et al. [8] extended these untargeted adversarial attacks to time series classification models and performed adversarial attacks on UCR time series datasets.

---

An example of a targeted adversarial attack is shown in Fig. 1. We are able to misguide a deep neural network model by adding imperceptible noise to the original sample such that the model predicts the class of our choice. All three plotted lines in Fig. 1 are visually indistinguishable but the deep learning model classify them as different classes according to our choice. Attackers can also modify the time series data of electricity consumption to hide the information of the device [9]. Similarly, time series data is used for preventive maintenance of equipment and to find health of the machine [3, 10]. Attackers can modify time series data in order to hide abnormal failure of an equipment in a safety critical application.

Goodfellow et al. [11] argued that neural networks are vulnerable to adversarial attacks not because of overfitting and non-linearity of the decision boundary but because of high dimensionality of the input space and local linearity of the model. If it is due to overfitting then models of different architecture will not make errors on the same adversarial example, which is not the case. Mustafa et al. [12] suggested that these adversarial examples exist because samples corresponding to different classes lie in close proximity in the learnt feature space and suggested that adversarial defense can be done by forcing the features of each class to lie inside a convex polytope which is maximally separated from the polytopes of the other classes. Dezfooli et al. [13] proposed universal adversarial perturbation for images and showed that same perturbation can be used to fool multiple images in a given dataset. Our findings show the existence of universal adversarial perturbation for time series data similar to image data. These perturbations are universal in the sense that the same perturbation vector can be used to fool different input samples. The contribution of our work is summarized as follows:

- We show the existence of targeted, untargeted, and universal adversarial perturbations for time series data and propose methods for finding these adversarial perturbations for deep learning based time series classification models.
- We showcase the generalization property of universal attacks i.e. attacks generated using only a fraction of the training samples are also able to misguide the classification models with a high success rate.
- We propose adversarial training based adversarial defense for deep learning based time series classification models.
- We demonstrate that models trained adversarially using Fast gradient sign method (FGSM), a single step attack, are able to defend against FGSM as well as Basic iterative method (BIM), a well-known iterative attack.

## II. BACKGROUND

In this section, we formally define the mathematical definitions of the time series classification problem, untargeted, targeted and universal attacks and adversarial training:

- $X^i \in R^T$ is the $i^{th}$ sample of the dataset $X$ and T is the sequence length of the sample
- $Y^i \in [0, K-1]$ where $Y^i$ is the true class for $i^{th}$ sample
- $K$ is the number of unique classes of $X$
- $X^i_{adv}$ is the adversarial sample corresponding to $X^i$
- $Y^i_T$ is the target class corresponding to $X^i$ which an attacker wants the model to predict
- $\varepsilon_{max} \in R$ is the upper bound of $L_\infty$ norm of the allowed perturbation for $X$
- $\varepsilon^i_{max} \in R^T$ is the upper bound of $L_\infty$ norm of the allowed perturbation for $X^i$
- $f(.) : R^T \to R^K$ is any deep learning model
- $\hat{Y}^i$ is the class predicted by $f(.)$ corresponding to $X^i$
- $L(f, X^i, Y^i)$ and $L_T(f, X^i, Y^i_T)$ are the losses corresponding to the sample $X^i$, deep learning model $f(.)$ for untargeted and targeted attacks respectively
- $N$ is the number of steps in BIM
- $\alpha \in R$ is the small step size for BIM
- $R_{fooling} \in R$ is the desired fooling ratio i.e. fraction of samples fooled from a given dataset
- $Epoch_{fool}$ is the maximum number of epochs to run to achieve $R_{fooling}$
- $U \in R^T$ is the universal adversarial perturbation for the dataset X
- $X^i_{FGSM, \varepsilon_{max}} \in R^T$ is the adversarial sample corresponding to $X^i$ using FGSM
- $Error(.)$ : computes misclassification error ratio of a dataset.

### A. Fast Gradient Sign Method (FGSM)

Goodfellow et al. [11] proposed FGSM attack, a single step attack, on images. It generates the adversarial sample by adding a small perturbation in $X^i$ in the direction of the sign of 'gradient of loss' w.r.t. input. Alternatively, the adversarial sample for an untargeted attack is obtained by

$$X^i_{adv} = X^i + \varepsilon^i . sign(\nabla_X L(f, X^i, Y^i)) \quad (1)$$

Adversarial sample for targeted attack where an attacker misguides the model to predict a target class $Y^i_{target}$ corresponding to $X^i$ is obtained by

$$L_T = -L \quad (2)$$
$$X^i_{adv} = X^i + \varepsilon^i . sign\left(\nabla_X L_T(f, X^i, Y^i_T)\right) \quad (3)$$

$L_T$ is equal to the negative of $L$. In case of targeted attacks, we minimize the loss between the predicted class and the target class whereas in case of untargeted attack we maximize the loss between the predicted class and the true class.

### B. Basic Iterative Method (BIM)

Kurakin et al. [7] suggested that applying FGSM iteratively on a sample by taking a smaller step size results in a stronger adversarial sample. After each iteration, the output is clipped to ensure that the adversarial sample lies within the $\varepsilon$-neighborhood of the original input $X^i$.

**Algorithm 1**: Computing universal adversarial perturbation

**Input:** $X, Y, f, \varepsilon_{max}, R_{fooling}, Epoch_{fool}$
**Output:** $U$
1: $U \leftarrow 0$
2: $Iter \leftarrow 0$
3: **while** $Error(X + U) < R_{fooling}$ And $Iter < Epoch_{fool}$ **do**
4:   **for** each sample $X^i \in X$ **do**
5:     **if** $\hat{Y}^i = Y^i$
6:       **if** $f(X^i_{FGSM,\varepsilon_{max}}) \neq Y^i$
7:         **if** $f(X^i + U) = Y^i$
8:           **Solve for optimum $r$ in the direction of** *gradient of loss* **w.r.t.** $X^i$ ($\nabla_X L(f, X^i, Y^i)$)
            $\Delta U = \underset{r}{\arg\min} \|r\|$ s.t. $f(X^i + U + r) \neq Y^i$
9:           Take projection of $U + \Delta U$ on $\varepsilon_{max}$ infinity ball
            $U \leftarrow Proj_{\varepsilon_{max}}(U + \Delta U)$
10:         **end if**
11:       **end if**
12:     **end if**
13:   **end for**
14:   $Iter \mathrel{+}= 1$
15:   Shuffle $X$
16: **end while**

---

Untargeted adversarial sample $X^i_{adv}$ is calculated using equations (4) and (5).

$$X^i_{adv,0} = X^i \quad (4)$$

$$X^i_{adv,N} = \min\{X^i + \varepsilon^i_{max}, \max\{X^i_{adv,N-1} + \alpha \cdot sign(\nabla_X L(X^i_{adv,N-1}, Y^i)), X^i - \varepsilon^i_{max}\}\} \quad (5)$$

Similarly, targeted adversarial sample $X^i_{adv,T}$ is calculated using equations (6) and (7).

$$X^i_{adv,T,0} = X^i \quad (6)$$

$$X^i_{adv,T,N} = \min\{X^i + \varepsilon^i_{max}, \max\{X^i_{adv,T,N-1} + \alpha \cdot sign(\nabla_X L_T(f, X^i_{adv,T,N-1}, Y^i_T)), X^i - \varepsilon^i_{max}\}\} \quad (7)$$

### C. Universal Adversarial Perturbation

In this section we mathematically define the notion of universal adversarial perturbations and a method to find such perturbations. We define a universal adversarial perturbation U for a given dataset X, such that it fools most of the samples chosen from the input distribution µ of $X$.

$$f(X^i + U) \neq Y^i \text{ for } X^i \sim \mu \quad (8)$$

Fig. 2 shows the geometric representation of universal adversarial perturbation represented by U [13]. We calculate universal adversarial perturbation that satisfies two conditions. Firstly, the infinity norm of the perturbation is less than or equal to $\varepsilon_{max}$. Secondly, it achieves the desired fooling ratio on a given dataset. Mathematically

$$\|U\|_\infty \leq \varepsilon_{max} \quad (9)$$

$$\underset{X^i \sim \mu}{P}(f(X^i + U) \neq Y^i) \geq R_{fooling} \quad (10)$$

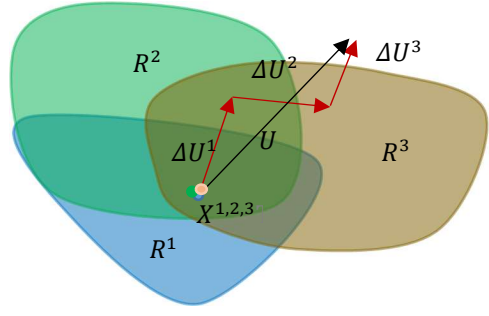

Fig. 2 Geometric representation of universal adversarial perturbation. $X^{1,2,3}$ represents the superimposition of samples, $X^1$, $X^2$ and $X^3$. $R^i$ is the region for correct class and $\Delta U^i$ is the additional adversarial perturbation in the direction of gradient of loss w.r.t. $X^i$ for i$^{th}$ sample. U is the universal adversarial perturbation for $X$

The method for finding such universal adversarial perturbations is described in algorithm 1.

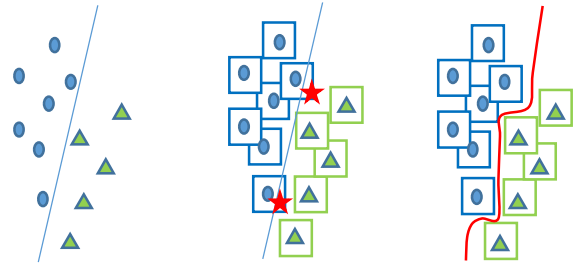

Fig. 3 Visualization of decision boundary (a) Left: Normal classifier (b) Middle: Stars indicate adversarial examples within rectangular $\varepsilon_{max}$ infinity balls (c) Right: Decision boundary of adversarially trained classifier

As mentioned in algorithm 1, $U$ is calculated by iterating over all samples in $X$ while neglecting those few samples for which predicted class is wrong as in step 5. We also neglect those "hard" samples for which FGSM attack is unsuccessful as in step 6. For each of the remaining samples, $\Delta U$ having minimum norm is computed by exploring different step sizes in the direction of the gradient of loss with respect to each sample. Further, projection of $U + \Delta U$ is taken on $\varepsilon_{max}$ infinity ball to ensure that the constraint $\|U\|_\infty \leq \varepsilon_{max}$ is satisfied. It is worth mentioning that there exists many universal adversarial perturbations for a given dataset. These are obtained by choosing different initial seeds for random shuffling of the dataset $X$.

Adversarial training is a strategy for adversarial defense. Algorithm 2 explains the method for adversarial training.

---

**Algorithm 2:** Adversarial training for each batch of data

1: for each batch **do**
2:   calculate $X_{adv}$ using BIM or FGSM
    *#For each sample there can be multiple adversarial*
    *#samples*
3:   concatenate $X, X_{adv}$
4:   calculate loss
5:   Update weights using backpropagation

Table 1: Summary of untargeted attacks, targeted attacks, universal adversarial attacks and adversarial training on UCR datasets

| Dataset Name | Original Accuracy of models (in %) | Accuracy of models after untargeted attacks (in %) | | | Success rate of targeted attacks (in %) | | Accuracy of adversarially trained models after (in %) | |
|---|---|---|---|---|---|---|---|---|
| | | Universal Attack | FGSM Attack | BIM Attack | FGSM Attack | BIM Attack | FGSM Attack | BIM Attack |
| 50words | 73.2 | 24.4 | 9.2 | 0.7 | 10.4 | 47 | 50.5 | 40.9 |
| Adiac | 83.1 | 3.1 | 2.8 | 0.5 | 2.7 | 15.1 | 67 | 14.1 |
| ArrowHead | 85.1 | 38.9 | 26.3 | 3.4 | 46.3 | 92.6 | 67.4 | 54.9 |
| Beef | 76.7 | 20 | 13.3 | 0 | 34.2 | 84.2 | 43.3 | 30 |
| Car | 93.3 | 23.3 | 21.7 | 0 | 26.1 | 76.1 | 68.3 | 25 |
| CBF | 98.9 | 98.3 | 85 | 83.7 | 7.8 | 9 | 95 | 95 |
| ChlorineConcentration | 83.5 | 37.1 | 1.6 | 0.9 | 95 | 98.7 | 87.9 | 87.8 |
| CinC_ECG_torso | 83.8 | 27.5 | 25.4 | 19.6 | 25.1 | 44.1 | 65.9 | 60.8 |
| Cricket_X | 79 | 75.1 | 26.2 | 10.5 | 24.1 | 50.4 | 59.2 | 57.4 |
| Cricket_Y | 80.5 | 52.6 | 14.6 | 2.6 | 24.5 | 64.7 | 57.2 | 52.8 |
| Cricket_Z | 81.5 | 63.8 | 17.9 | 5.9 | 28.7 | 55.3 | 55.9 | 52.8 |
| DiatomSizeReduction | 30.1 | 30.1 | 30.1 | 3.9 | 23.3 | 43.5 | 92.2 | 89.9 |
| DistalPhalanxOutlineAgeGroup | 79.8 | 12.3 | 2.3 | 0 | 75.8 | 98.8 | 90.5 | 93.5 |
| DistalPhalanxTW | 74.7 | 16.3 | 0.3 | 0 | 31.9 | 74.1 | 82.8 | 66.8 |
| ECG5000 | 93.5 | 87.5 | 73.2 | 32 | 9 | 24.1 | 92.1 | 88 |
| ElectricDevices | 73.5 | 48.2 | 43.8 | 21.7 | 12.3 | 34 | 61.5 | 43.8 |
| FaceAll | 85.5 | 85.5 | 69.1 | 65.1 | 5.4 | 7.9 | 72.9 | 71.6 |
| FaceFour | 95.5 | 95.5 | 70.5 | 42 | 20.5 | 41.3 | 54.5 | 54.5 |
| FacesUCR | 95.3 | 95.3 | 77.4 | 74 | 3.9 | 4.8 | 79.8 | 78.1 |
| FISH | 97.7 | 13.1 | 12.6 | 3.4 | 14.6 | 75.9 | 86.9 | 54.9 |
| Haptics | 51.6 | 18.8 | 18.8 | 0.6 | 20.4 | 78 | 27.9 | 15.3 |
| InlineSkate | 37.8 | 15.1 | 13.6 | 1.6 | 16.1 | 50.5 | 28.4 | 6 |
| InsectWingbeatSound | 50.6 | 20.7 | 9.2 | 0.5 | 16 | 65.2 | 39.9 | 21.9 |
| LargeKitchenAppliances | 90.4 | 90.7 | 69.9 | 60.3 | 21.2 | 29.7 | 86.7 | 84.3 |
| Lighting7 | 78.1 | 78.1 | 31.5 | 16.4 | 23.1 | 33.8 | 39.7 | 30.1 |
| MALLAT | 96.6 | 19.8 | 30.7 | 1.4 | 26.2 | 71.7 | 46.9 | 14.8 |
| Meat | 98.3 | 33.3 | 33.3 | 0 | 33.3 | 72.5 | 98.3 | 6.7 |
| MedicalImages | 76.2 | 67.4 | 40.1 | 11.6 | 12.2 | 31.5 | 54.1 | 42.5 |
| MiddlePhalanxOutlineAgeGroup | 74.2 | 33.8 | 49.3 | 0 | 31.3 | 97.8 | 86.3 | 58.8 |
| MiddlePhalanxTW | 60.9 | 11.5 | 4 | 0 | 34.3 | 93.2 | 67.7 | 62.2 |
| NonInvasiveFatalECG_Thorax1 | 94.6 | 3.7 | 5.3 | 0.1 | 2.9 | 62.1 | 86.7 | 37.1 |
| NonInvasiveFatalECG_Thorax2 | 94.4 | 3.3 | 5.1 | 0 | 3.4 | 27.7 | 85.3 | 65.2 |
| OliveOil | 86.7 | 33.3 | 20.0 | 0 | 28.9 | 88.9 | 90 | 30 |
| OSULeaf | 97.9 | 25.6 | 15.7 | 0 | 21.3 | 86.5 | 86 | 80.2 |
| Phoneme | 33.3 | 8.8 | 4.9 | 1.2 | 11.9 | 27.5 | 7.8 | 4 |
| Plane | 100 | 100 | 81 | 56.2 | 7.6 | 16.3 | 100 | 100 |
| ProximalPhalanxOutlineAgeGroup | 83.9 | 7.8 | 42.4 | 0 | 52 | 96.6 | 91.2 | 77.1 |
| ProximalPhalanxTW | 77.8 | 5.3 | 8.8 | 0 | 23.9 | 68.2 | 80.5 | 67.8 |
| RefrigerationDevices | 51.7 | 53.9 | 4 | 1.9 | 82 | 90.9 | 52 | 49.9 |
| ScreenType | 60.8 | 46.1 | 11.2 | 2.9 | 69.2 | 90.3 | 46.7 | 30.9 |
| ShapesAll | 91.7 | 8.5 | 4.8 | 0 | 2.4 | 20.2 | 75.7 | 73.7 |
| SmallKitchenAppliances | 78.9 | 33.3 | 41.6 | 14.4 | 30.7 | 58.5 | 81.1 | 52 |
| StarLightCurves | 97.2 | 58.5 | 58.8 | 57.7 | 21 | 29.2 | 97.7 | 77.6 |
| SwedishLeaf | 95.4 | 56.2 | 27.4 | 12.2 | 14.3 | 43.6 | 89.4 | 78.2 |
| Symbols | 92.7 | 74.9 | 31.2 | 7.3 | 18.9 | 35.2 | 70.3 | 63.3 |
| synthetic_control | 100 | 99.7 | 94.3 | 94 | 1.1 | 1.2 | 95.3 | 95.3 |
| Trace | 100 | 34 | 58 | 52 | 14.7 | 23.7 | 100 | 99 |
| Two_Patterns | 100 | 100 | 98.2 | 96.7 | 0.6 | 1.1 | 100 | 100 |
| uWaveGestureLibrary_X | 78 | 16.1 | 26.2 | 0.9 | 19.6 | 77.2 | 77.4 | 48 |
| uWaveGestureLibrary_Y | 66.7 | 31.8 | 20.1 | 0.2 | 18.9 | 78.5 | 72.4 | 48.5 |
| uWaveGestureLibrary_Z | 75 | 33.4 | 30.3 | 1 | 16.3 | 80.6 | 60.6 | 34.2 |
| UWaveGestureLibraryAll | 86.2 | 35.6 | 17.1 | 0 | 31.1 | 96.8 | 66.7 | 40 |
| WordsSynonyms | 62.5 | 44.5 | 3.8 | 0.6 | 18.7 | 59.7 | 43.3 | 36.7 |
| Worms | 64.6 | 53 | 19.9 | 3.3 | 37.7 | 87.6 | 61.3 | 60.2 |

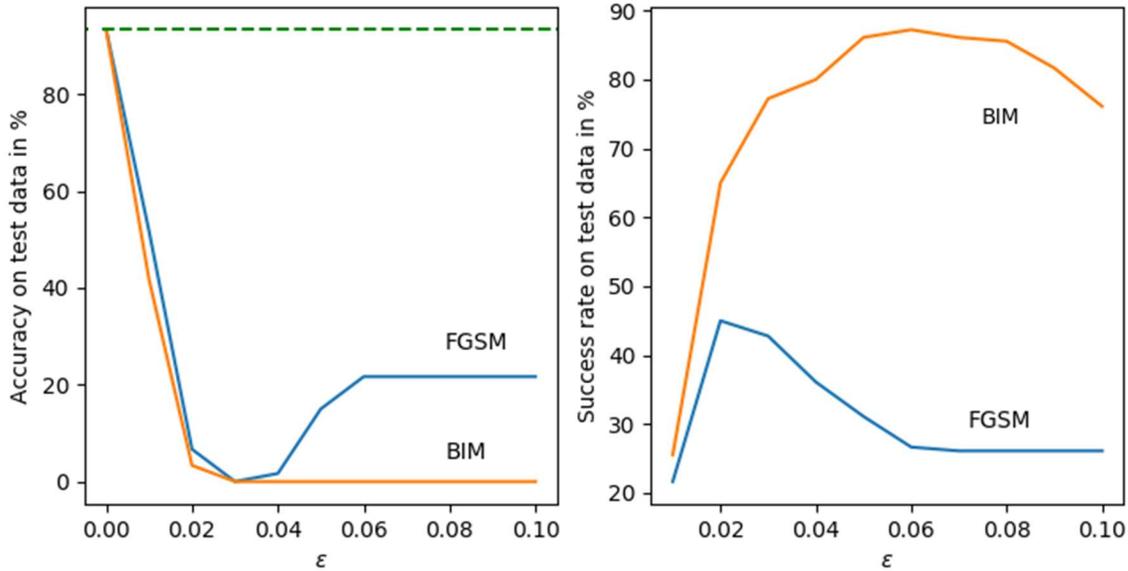

Fig. 4 (a) (Left) Effect of varying epsilon on model accuracy after FGSM and BIM based untargeted attacks for 'Car' dataset. Dashed line represents the original accuracy (b) (Right) Effect of varying epsilon on success rate of targeted FGSM and BIM attacks for 'Car' dataset

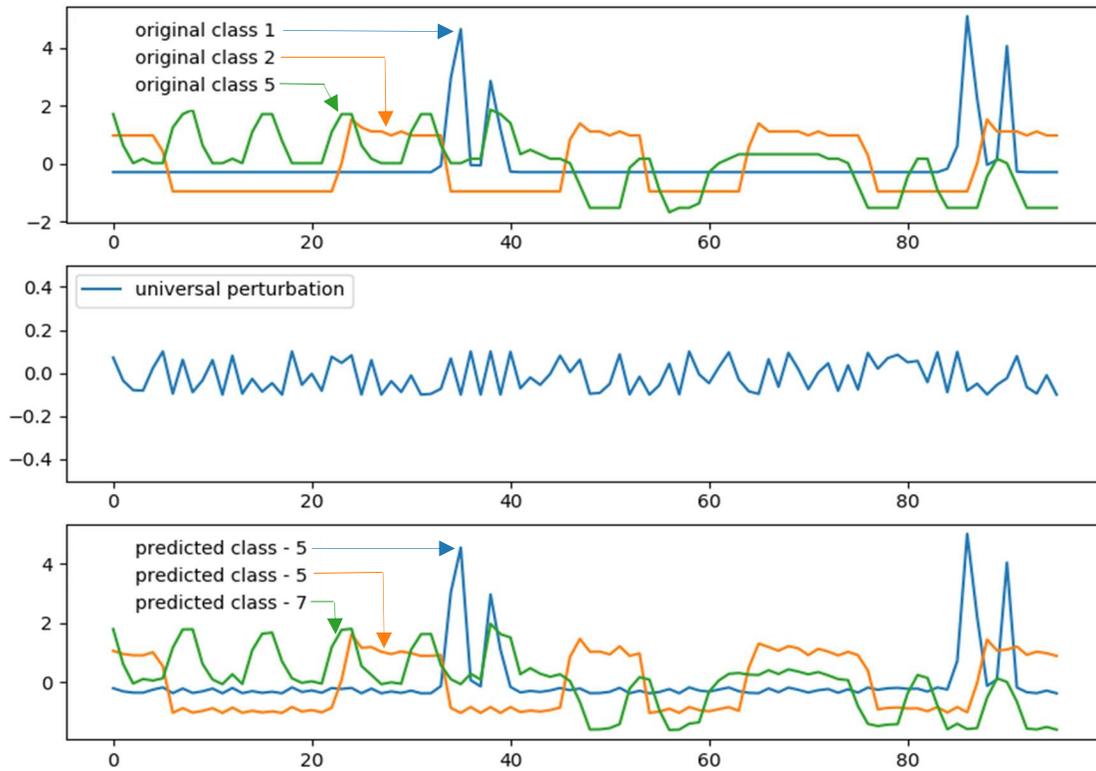

Fig. 5 (a) (Upper) Three samples from 'Electric devices' dataset having different classes (b) (Middle) Universal adversarial perturbation for the 'Electric devices' dataset (c) (Lower) Corresponding samples after adding universal adversarial perturbation

During adversarial training, adversarial samples are used for augmenting training data. Szegedy et al. [4] showed that adversarial training i.e. training model using original as well as corresponding adversarial samples acts as a regularizer for image data. An illustration of decision boundary in case of normal training as well as adversarial training is shown in Fig 3 [14]. Adversarial training tries to change the decision boundary in such a way that the $\varepsilon_{max}$ infinity ball around each data point does not cross the decision boundary and hence, it becomes more difficult to find any successful imperceptible adversarial perturbation. As mentioned in algorithm 2, each batch of training data comprises of original sample along with

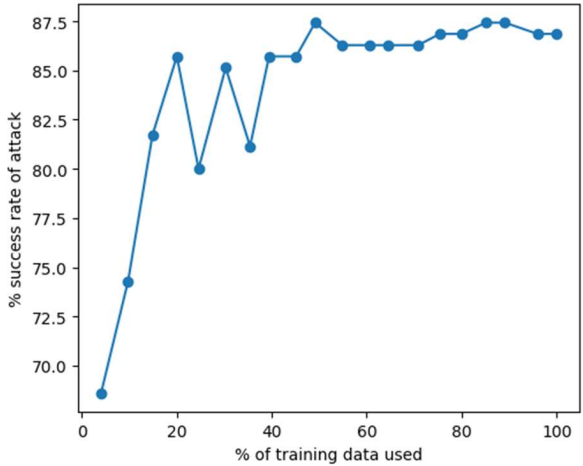 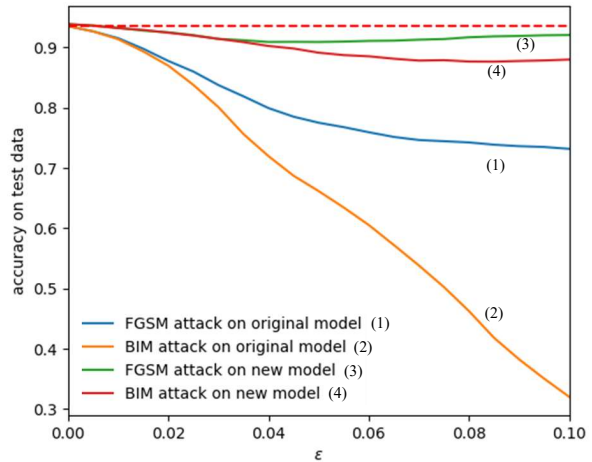

Fig. 6 (a) (Left) Variation of success rate of universal adversarial attacks with percentage of training data used for generating them. The figure is generated using 'Fish' dataset (b) (Right) Variation of accuracy of original model and adversarially trained model with perturbation level ε. The dashed line represents the accuracy of the original model. The figure is generated using 'ECG5000' dataset their corresponding one or multiple adversarial counterparts.

### III. RESULTS AND EXPERIMENTS

In this section, we discuss the various numerical experiments conducted and summarize the findings. We have performed all our experiments on the 54 multiclass univariate time series datasets taken from the UCR archive [15]. UCR time series dataset is a collection of diverse real life use cases covering food quality and safety, electrical equipment, vehicle sensors, healthcare etc. Hence, our study may be applicable across industries.

For time series classification we have used state of the art deep learning models based on resnet architecture [16]. Our model architecture consists of 3 residual blocks where there are 3 convolutional layers in each block [8, 17]. The final convolutional layer output is fed to a Global Average Pooling (GAP) layer which averages out the activations in the time direction and finally the whole network outputs a probability distribution over multiple classes through a softmax layer. In all the experiments, the maximum value of allowed pointwise perturbation, $\varepsilon_{max}$ is taken as 0.1 unless specified otherwise. All results shown in this work are on test datasets.

#### A. Untargeted and Targeted Adversarial Attack

We have performed untargeted and targeted adversarial attacks on 54 multiclass datasets using both single step (FGSM) and multistep (BIM) attacks. For BIM we have used the values of $N$ and $\alpha$ as 10 and $\varepsilon_{max}/2$ respectively. Model accuracy after untargeted attacks and success rate of targeted attacks are shown in Table 1. Success rate of targeted attack on dataset X with N classes and M samples is defined as the percentage of targeted attacks successfully achieved out of all possible $(N-1)*M$ combinations.

It can be observed from Table 1 that in case of untargeted attacks, iterative attacks (BIM) are able to reduce the model accuracy more than the single step attacks (FGSM) for all the datasets. Similarly, in case of targeted attacks we see that BIM has higher success rate than FGSM. For 'Car' dataset, BIM and FGSM attacks reduce model accuracy from 93.3% to 0% and 21.7% respectively. This is reflected both in Table 1 and Fig. 4(a). In Fig. 4 we have shown effect of varying $\varepsilon$ on model accuracy in case of untargeted attacks and on success rate in case of targeted attacks. Although plots are shown only for the 'Car' dataset we have found similar trends in most of the other datasets as well. As shown in Fig. 4(a), classification accuracy on test data in untargeted attacks decreases with the increase in ε till some optimum values of ε for both FGSM and BIM attacks. However, beyond the optimum value of ε, classification accuracy may increase a little or remain almost constant as shown in Fig. 4(a). In case of targeted attacks success rate of both FGSM and BIM initially increases with increase in ε and then decreases or remains constant beyond some optimum value of ε as shown in Fig. 4(b).

Almost always iterative attacks (BIM) have better performance than single step attacks (FGSM). Performance of both BIM and FGSM are better at large values of $\varepsilon_{max}$ as compared to small values of $\varepsilon_{max}$. Intuitively, at large values of maximum allowable perturbation level i.e., $\varepsilon_{max}$ there exists larger space for finding adversarial samples and hence, attacks are more powerful. However, at very large $\varepsilon_{max}$ the input time series signal itself gets distorted and the additional adversarial perturbation no longer remains 'imperceptible'. As a result, the classification model predicts any random class. We think that these two counteracting effects are giving rise to optimum values of $\varepsilon_{max}$ as shown in Fig. 4.

As shown in Table 1, in case of untargeted attack for all the datasets both FGSM and BIM are able to reduce model accuracy except the 'DiatomSizeReduction' dataset for which FGSM did not work. 'DiatomSizeReduction' dataset has the smallest training dataset size with only 16 training samples and low original classification accuracy of 30.1% due to overfitting. We have also found that both FGSM and BIM attacks perform poorly on some simulated datasets like 'Two_Patterns' and 'synthetic_control'. These simulated datasets have robust classifiers with high original test accuracies. Apart from this, 'synthetic_control' dataset has the smallest sequence length making it hard to attack. For these simulated datasets, we have found low reduction in accuracies during untargeted attack and low success rate during targeted attack.

*B. Universal adversarial perturbation*

We have shown the existence of universal adversarial perturbations for time series datasets. For a given dataset, there exist multiple such perturbations. Existence of universal adversarial perturbations increases the threat to real life systems based on deep learning as it signifies that there exist directions in the input data space which if exploited can fool most of the samples of a given dataset. As shown in Fig. 5, the universal adversarial perturbation for 'Electric devices' dataset is able to fool 3 different samples from different classes. There exists multiple universal adversarial perturbations for a given dataset which can be computed by choosing different initial seeds for the random shuffling of the training dataset while using algorithm 1. We also show that these universal adversarial perturbations have very good generalization property i.e. universal perturbations computed using only a subset of the training dataset are also able to fool the model and have good success rate on test dataset. The concept of "success rate" in case of universal adversarial attack is a little bit different than that of targeted attacks using FGSM and BIM. Here success rate is the percentage of samples incorrectly predicted by the model after universal adversarial attack.

Fig. 6(a) shows the variation of 'success rate of universal adversarial attack' for 'Fish' dataset w.r.t. different percentages of training data used for generating the universal adversarial perturbations. As shown in Fig. 6(a), less than 10 percent of training samples are required to achieve 70 percent success rate in the case of 'Fish' dataset. Success rate of universal adversarial attack increases with the increase in percent of training data used for computing them and becomes almost constant after some percentage of training data. In Table 1 we have shown classification accuracy after universal adversarial attack for all the 54 multiclass datasets. Universal adversarial attack was able to reduce model classification accuracy for 45 datasets out of 54 multiclass datasets. Similar to FGSM, universal adversarial attack was also not able to reduce model accuracy for the dataset 'DiatomSizeReduction'. Apart from this, universal adversarial attack performed very poorly on the simulated datasets like 'Two_Patterns', 'synthetic_control', 'CBF' etc. In our experiments, we took the values of $R_{fooling}$ and $Epoch_{fool}$ as 0.9 and 10 respectively.

*C. Adversarial Training*

Here we explored the idea of increasing the robustness of any deep learning model against adversarial attacks by incorporating adversarial samples into the training data during training. In all relevant experiments we have directly used the original pretrained model, referred to as "original model" in Fig. 6(b), to generate the adversarial samples. For each sample in the training dataset, we have calculated a corresponding adversarial sample using untargeted FGSM for later use. During training, the trainable model was initialized using the original pretrained model. In each minibatch, original training samples along with their corresponding already computed adversarial samples were fed to the network. The initial learning rate and the minimum learning rate was taken as 0.0005 and 0.0001 respectively. We reduced learning rate by a factor of 0.5 whenever training loss did not reduce for 50 consecutive epochs. We ran training for 1500 epochs to get the final updated model which is referred to as "new model" in Fig. 6(b).

In Fig. 6(b) we have studied the effect of adversarial samples generated for different $\varepsilon$ values on accuracies of both original and new models. Fig. 6(b) shows results for 'ECG5000' dataset. It is clear that our implementation of adversarial training using only one attack method (FGSM) is able to impart defense in the new model against both FGSM and BIM attacks. Besides, models adversarially trained for one value of $\varepsilon_{max}$ seems to have impressive defense against attacks for other values of $\varepsilon$ where $\varepsilon$ is less than $\varepsilon_{max}$. Intuitively, for large values of $\varepsilon_{max}$ the adversarial attacks are more powerful. So, models adversarially trained using large values of $\varepsilon_{max}$ are able to defend against weaker attacks corresponding to smaller values of $\varepsilon$. In the last two columns of Table 1, we have reported the accuracies of our 'models adversarially trained using FGSM' on test data samples adversarially attacked by FGSM and BIM respectively.

CONCLUSION

In this work, we have shown that there exist untargeted, targeted and universal adversarial perturbations for time series datasets. We proposed a novel algorithm for finding universal adversarial perturbations and performed a detailed study on all 54 multiclass UCR time series datasets. Since, powerful universal adversarial perturbations can be computed easily, it poses serious threat to real life applications based on deep learning models. Iterative attacks (BIM) are found to be more powerful than single step attacks (FGSM) for untargeted as well as targeted attacks. We demonstrated that adversarial training increases the robustness of deep learning models against adversarial attacks. Also, 'adversarially trained models' trained using FGSM (a single step attack) samples are able to defend against FGSM as well as BIM, an iterative attack. In future, effect of adversarial training by using multiple adversarial attacks like FGSM, BIM, Universal adversarial attack etc. can be studied.

ACKNOWLEDGMENT

We would like to thank the authors of "Adversarial Attacks on Deep Neural Networks for Time Series Classification, 2019" [8] and UCR data archive [15] for providing the pre-trained models and time series data respectively. We have used these datasets and pre-trained models in our study.